# A new bio-inspired method for remote sensing imagery classification

Amghar Yasmina Teldja*, Fizazi Hadria[†]

The problem of supervised classification of the satellite image is considered to be the task of grouping pixels into a number of homogeneous regions in space intensity. This paper proposes a novel approach that combines a radial basic function clustering network with a growing neural gas include utility factor classifier to yield improved solutions, obtained with previous networks. The double objective technique is first used to the development of a method to perform the satellite images classification, and finally, the implementation to address the issue of the number of nodes in the hidden layer of the classic Radial Basis functions network. Results demonstrating the effectiveness of the proposed technique are provided for numeric remote sensing imagery. Moreover, the remotely sensed image of Oran city in Algeria has been classified using the proposed technique to establish its utility.

## I.  INTRODUCTION

For remote sensing applications, classification is an important task which partitions the pixels in the images into homogeneous regions, each of which corresponds to some particular landcover type. The problem of pixel classification is often posed as clustering in the intensity space.

Some natural events are able to implement original heuristics to solve problems for which it is difficult to find out solutions deterministically by classical algorithms. Further, these heuristics are robust; even with the failure of a constructive component of the heuristic can't affect the heuristic at all.

The source of bio-inspired computing is the behavior of social insects: a population of simple agents interacting and communicating indirectly through their environment is a massively parallel algorithm for solving a given task, such as the foraging, the crowd, the task division and prey capture …

Indeed, for some years, lot of studies revealed the effectiveness of the stochastic approach based on ant colony for solving various problems; as the combinatorial optimization problems - such as such as robotics [1] and the traveling salesman problem [2] - and algorithms based on ant - where the main constitute the metaheuristic of the ant colony optimization ACO - making incrementally configurations [3] [4].

Ants are able to solve complex problems collectively, as finding the shortest path between two points in a broken environment. The collective capacity of the ants to find shortest path is mainly due to the fact that more the path is short, more quickly the ant returns to the nest by this road, by redepositing the pheromone, and more ants are attracted on this path and reinforce it. [2]

The classification forms also part of these problems in which the ants suggest very interesting heuristics for processing specialists. Based on existing work, we contribute in this work to the study of classifiable ants as seen from the knowledge discovery, with the goal of solving real problems. In such problems, we consider that a domain expert has collected a data set and he would like to see himself proposing a partition of his data in relevant classes. [1]

The artificial immune systems form a rather recent research sector compared with other models of data-processing calculation taking as a starting point the biology to find solutions with various encountered problems. These are proper adaptive defense systems, able to create a very large variety of cells and molecules capable to recognize and eliminate specifically practically unlimited number foreign invaders. These cells and these molecules intervene together in a dynamic network precisely adaptable, whose complexity competes with that of the nervous system. [5]

This work is placed within the framework of a supervised classification; supposing that the number and the parameters of the classes are known. The application is carried out using a neurons network: the radial basis function network including the growing neural gas with utility factor (RBF-GNGU or RBFU) [6], having for task to treat the image pixels according to the various components categories the western area of Oran city. In effect, this area consist on many components (vegetation, water,…), which is the source of a great variability of reflectance, which is translated at the same time between various pixels and inside the same pixel whose reflectance can then be a mixture of several other components.

## II. ANT COLONY

Biologists have long been intrigued by the behavior of insect colonies: ants, bees, termites …. Each individual of the colony is a priori independent and is not supervised in one path or another. It is helped by the community in its development and in return it helps the proper functioning of this one. The colony is self-controlled through simple mechanisms to study. [7]

Walking from the nest to the food source and vice versa, the ants deposit in passing on the ground an odorous substance called "pheromone". So this substance allows creating a chemical trail, and other ants can detect pheromones through sensors on their antennae, and follow the same traversed path.

Pheromones have a role as a path marker: when the ants choose their path, they tend to choose the track with the highest concentration of pheromone. This allows them to find their path back to their nest upon return. On the other hand, smells can be used by other ants to find food sources found by their peers. [7]

Consequently, this behavior allows the emergence of shortest paths, provided that the pheromone trails are used by an entire colony of ants (see Fig. 1 and Fig. 2).

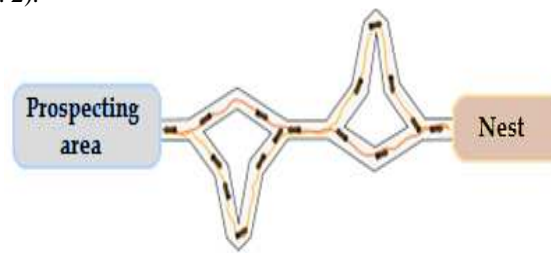

Fig. 1. Behavior at beginning

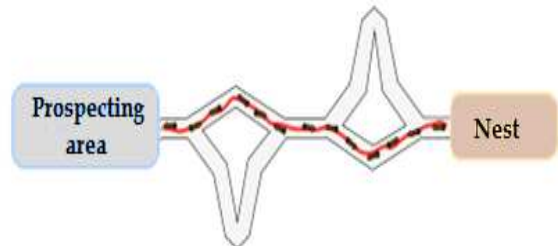

Fig. 2. Behavior after 10 min

This behavior allows finding the shortest path to food when the pheromone trails are used by the entire colony. In other words, when multiple marked paths are at the disposal of an ant, it can find the shortest path to its destination. This essential observation is the base of all the methods which we will develop further [7].

Ants lay pheromone on the path to the food source and back to the nest. Initially, the choice is random but quickly the short arm becomes most pronounced because the ants that use it arrive faster to the nest and will be statistically more likely to borrow when they return to the food source.

From this behavior is born the ant colony optimization metaheuristic (Ant Colony Optimization). ACO is a class of algorithms, whose first member, called Ant System, was initially proposed by Colorni, Dorigo and Maniezzo [8]. Several algorithms based on or inspired by the ant colony optimization metaheuristic have been proposed to tackle continuous optimization problems [9]. The first Ant System was proposed in the earlier nineties, and since then several studies were performed to apply this paradigm in real problems. Several researchers have explored the idea of applying it to image processing. [10]

Some features of the behavior of artificial ants are inspired by nature, thus, the artificial ants deposit pheromone on the arcs of the graph which they borrows, and choose their path randomly depending on the quantity of pheromone deposited on the incident arcs. In addition, the quantity of pheromone is gradually reduced, simulating an evaporation phenomenon avoiding premature convergence. [2]

Artificial ants have also other characteristics, which do not find their counterpart in nature. In particular, they may have a memory, which allows them to keep track of their past actions. In most cases, the ants deposit a pheromone trail only after performing a full path and not incrementally, progressively when they move.

Here, the work is carried around the modeling of a particular type of ants: the "*Pachycondyla apicalis*". This study stems from the work of Dominique Fresneau [11] on the original foraging strategy of this species of ponérines ant. So, we start by presenting the biological characteristics that could be exploited in terms of algorithmic modeling with the aim to apply for our optimization problem.

### A. *Pachycondyla apicalis* ant

*Pachycondyla apicalis* is a common and conspicuous insect in many Neotropical forests. Most observations and collection records are of single foragers on the ground or on low vegetation. This species has been observed nesting in rotting wood on or near the ground, in the ground, and in the root mass of large ficus trees within one meter of the ground.

In wet and dry forested lowlands, *Pachycondyla apicalis* is one of the most common and conspicuous ants. Foraging workers are extremely fast and run over the surface of trails in a nervous, erratic manner, with the antennae rapidly vibrating. Their behavior is reminiscent of compiled wasps.

Colonies are small, containing fewer than 200 workers [12], [13], and monogynous [14].

Foraging is done individually, without recruitment, and individual foragers over time show strong fidelity to a particular area. Tandem-running has been observed during nest relocation. Orientation is probably visual, test an optimal foraging model using Pachycondyla *apicalis*, concluding that foraging in the observed colonies is sub-optimal. A group of computer scientists have used the foraging behavior of *Pachycondyla apicalis* as a model for creating an internet search algorithm [15].

Our interest for these ants is that they use simple principles, of a global perspective and local levels, to find their prey. From their nest, they generally cover a given surface in the partitioning in individual hunting sites. [11]

### B. *Local behavior of the ant*

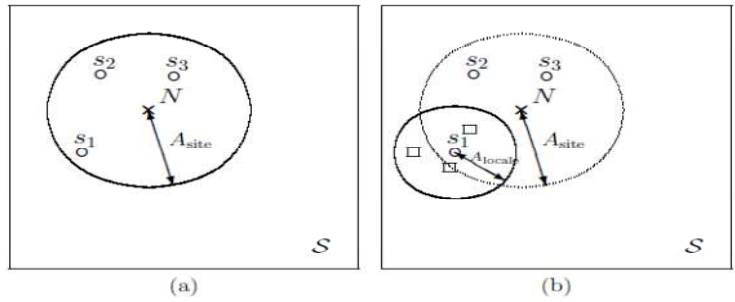

Fig. 3. **(a):** Search for hunting sites, **(b):** Local exploration around the site $s_1$

In figure 3.a, the ant constitute is a list of three hunting sites $s_1$, $s_2$ and $s_3$ in the vicinity of the nest $N$ with a maximum distance of $A_{site}$ from de nest. While in figure 3b, the squares represent the explorations conducted in the area radius $A_{local}$ around of $s_1$.

Initially, and every movement of the nest, each ant $a_i$ leaves the nest to provide a list of $p$ hunting sites which it stores (see fig. 3.a). A hunting site is a point of $S$ constructed by the operator $O_{explo}$ with amplitude $A_{site}(a_i)$ in the vicinity of $N$. The ant $a_i$ will then perform a local exploration around one of his hunting sites (see fig. 3b). [11]

Initially, when the interest of the sites is unknown, the ant selects a site $s$ randomly from the $p$ site that its disposal. The exploration is to build a local point $s'$ of S in the vicinity of $s$ due to the operator $O_{explo}$ with a amplitude $A_{local}(a_i)$. The ant $a_i$ capture a prey if the local exploration has to find a better value of $f$, which is to be had $f(s') < f(s)$. Improved of $f$ models so the capture of prey. Each time an ant is able to improve $f(s)$, it stores $s'$ instead of $s$ and the next exploration will take place in the local neighborhood of $s'$. If the local exploration is unsuccessful, the next exploration, the ant will choose a site randomly from the $p$ sites it has in memory. When a site has been explored more than $P_{local}$ times without having reported prey, it is definitely forgotten and will be replaced by a new site in the next release of the nest so the next iteration. The parameter $P_{local}$ represents a local patient. [11]

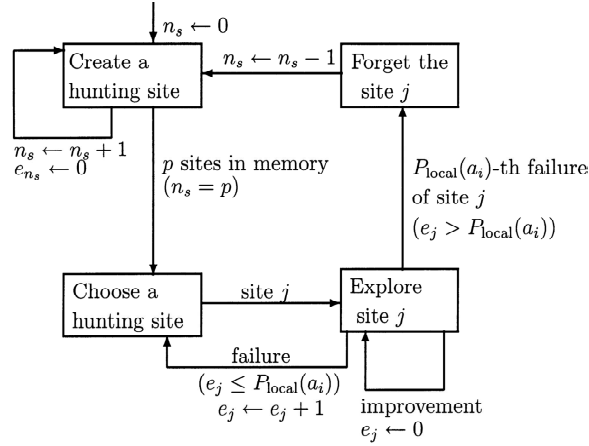

Fig. 4.  Automaton representing the individual behavior of an ant foraging

The automaton of figure 4 summarizes the individual behavior of forage. $n_s$ is the number of sites for hunting the ant stores at any given time. $e_j$ gives the number of successive failures encountered on the site memorized $s_j$. [11]

The main steps in the simulation of the Pachycondyla apicalis ant are given by the following algorithm:

*API()*
Randomly choose the initial location of the nest:
$N \leftarrow O_{rand}$
$T \leftarrow 0$         / * index of the iterations number * /
  **While** the stopping condition is not satisfied **do**
    **For every** $a_i \epsilon A$ **do**
      API-Foraging $(a_i)$
    **If** the nest must be moved **then**
      $N \leftarrow s^+$    / * Best solution reached * /
      Clear the memory of all ants
  $T \leftarrow T + 1$
**Return** $s^+$ and $f(s^+)$

The stop condition can be various [11]: either $s^+$ has not been improved since a number of iterations $(T_1)$, or $T$ has reached a limit value $(T_2)$ or a number $(T_3)$ of evaluations of $f$ solutions was reached.

### III.   IMMUNE SYSTEM

The biological immune system's basic functioning identifying antigens, and the affinity between receptor cells and the antigens, whether they are complete or decomposed, from there was create the "Artificial Immune System" or commonly known as " AIS".

To be used in computing, the different components of the biological immune system, and particularly immune cells and antigens, should be modeled as digital and date form.

For the modeled cells can recognize the "antigens", this concept of affinity is taken again and applied to the data-processing cells. The lymphocytes B and T have on their surface a great number of receivers, each one is able to recognize only one type of antigen and all the receivers of the same cell are identical.

In the AIS - which the application domain is the data processing - we'll not tell B or T cells or their respective receptors any more, but an antibody in general to facilitate modeling and representation. [5]

The selection concept is included in the AIS:
- *Positive selection:* it ensures that all the T cells leaving the thymus recognize the MHC (**M**ajor **H**istocompatibility **C**omplex) of the self.
- *Negative selection:* it ensures that all the T cell (T lymphocyte) recognizing too much the self cells as antigen doesn't leave the thymus in order to prevent autoimmune diseases.
- *Clonal selection (CS):* this is the theory explaining how the immune system interacts with the antigen. This theory is applicable to B and the T cell. The only difference is that the B cells undergo somatic hypermutation during proliferation in contrast to the T cells. The AIS are inspired by this theory, but when only B cells are able to mutate for optimize the immune response, only those cells interest us.

This optimization is the fact that B cells upon contact with antigen proliferate and give several clones and each clone is mutated. This mutation is used to find clones of the mother cell with a greater affinity with the antigen. In the successive mutations, there is the risk of being faced with a problem of local optimum which we can't find a better clone even in passing generations of cells. The human body avoids this problem by adding to these clones some newly created elements, which do no have any relation with the mother cell. This addition of "random" elements solves the problem by changing the new elements if they have a greater affinity with the antigen. These cells are presented as bit vectors of length $L$. [5]

Figure 5 shows the clonal selection algorithm, which proceeds as follows:

(1) Generate randomly a number of defined receptors,
(2) Select from these receptors the $n$ best,
(3) Clone the new population,
(4) Mutate the new population,
(5) Refilter the obtained population keeping only the best elements (memory cells),
(6) Replaced a portion of these cells by other detectors randomly regenerated.

The introduction of these new detectors avoids the local optima problem. Then we start the cycle again by making the best selection of $n$ elements [5].

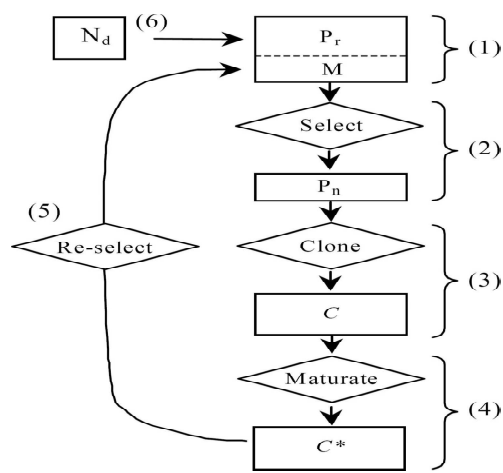

Fig. 5. Clonal selection algorithm

The mutation of the detectors is performed by changing one or more bits of the vector representing the cell by another value. The main differences between the clonal selection algorithms are the methods used in the random generation of detectors, the mutation and the affinity between the detectors and antigens. [5]

### A. Principle of clonal selection

When a B lymphocyte receptor (LB) identifies an antigen with a certain affinity, it was selected to proliferate. This growth will result in the production of a cell clone of a single type. Due of the mutation, the cells within a clone are similar but have slight differences that can recognize the antigen that triggered the immune system response.

This whole process of mutation and selection is known as the immune response maturation. The B lymphocyte with high antigenic affinities are selected to become memory cells with a long life. These memory cells predominate in future responses to the same antigen or similar.

Other important features of clonal selection appropriate from the viewpoint of the cell are:
1. An antigen is recognized by several B cell. The rate of proliferation of each cell B is proportional to its affinity with the antigen selected (higher is the affinity, greater is number of offspring (clones) product, and vice versa).
2. Inversely with the rate of proliferation, mutation suffered by each B cell during the reproduction is inversely proportional to the affinity of the receptor of B cells with antigen (higher is the affinity, smaller is the mutation, and vice versa).

### B. Clonal selection algorithm :

The clonal selection algorithms are often used in optimization applications seen as B cells become more affine to antigens, intrusion detection applications where we can not list all the undesirable elements and where elements are extremely varied. [5]

This whole process of mutation and selection is known clonal selection algorithm, originally known as the CSA (Clonal Selection Algorithm) or CLONALG (Clonal Algorithm) is inspired by the following elements of the clonal selection theory:

- Maintenance of a specific memory location.
- Selection and cloning of most stimulated antibodies.
- The death of unstimulated antibodies.
- Affinity maturation (mutation).
- Re-selection of clones proportional to the affinity of the antigen. Generation and maintenance of diversity.

  *a) Main elements of the clonal selection algorithm :*
- **Antigen:** in the AIS, the antigen means the problem and its constraints.
- **Antibodies:** the antibodies represent candidates of the problem.
- **Affinity antibody / antigen:** the reflection of the power of combination placed between the antigen and the antibodies.
- **Mutation (affinity maturation):** random change of the antibody value.
- **Memory cell (antibodies memory):** the memory cell represents the best antigen.

  *b) CLONALG algorithm:*

*1) Initialization:* in this initialization step, the algorithm generates a random antibody population with size *N*. Then the population is divided into two components, a memory antibodies section *m*, and a remaining group *r*.

*2) Generation:* the algorithm proceeds to the execution of a number of iterations to expose the system to all known antigens. One round of exposition or iteration is considered as a generation. The number of generations *G* performed by the system is predefined by the user.

- **Select of the antigen:** an antigen is selected randomly, without replacement (for de actual generation) of the population of antigens.
- **Exposure:** the system is exposed to the selected antigen. The affinity is calculated for all the antibodies directed against the antigen. The affinity is a measure of similarity, it depends on the problem. The Hamming distance is most often used in this case.
- **Selection:** a set of *n* antibodies is selected from the entire population of antibodies with the highest affinity for the antigen.
- **Cloning:** a set of selected antibodies is then cloned according to their affinity (founded classification).
- **Affinity maturation (mutation):** the clone (antigen duplicated) is then subjected to a process of affinity maturation to respond better to the antigen.
- **Clone exposition:** the clone is exposed to the antigen, and affinity measures are calculated.
- **Candidature:** one or several antibodies having the highest affinity with the clone are selected as a memory antibody for a replacement in *m*. If the affinity of the candidate memory cell is higher than the largest, then it is replaced.
- **Replacement:** the *d* individuals in the remaining antigen population with the lowest affinity are replaced by new random antibodies.

*3) Output:* after the achievement of the mode of formation, the memory component *m* of the antigen is considered as the solution of the algorithm. This solution must be the best individual or the collective association of all the antigens in the population.

The algorithm CLONALG applied to the pattern recognition use the following formulas to calculate the parameters:

- **Number of clones :** created from each antibody, it is calculated as follows:

$$nbrClones = \frac{\beta * N}{i}$$

Where $\beta$ is a clonal factor, $N$ is the size of the antibodies population, and $i$ is the antibody rank with $i \in [1, n]$.

The number of clones prepared for each antigen exposition to the system is calculated as follows:

$$N_C = \sum_{i=1}^{n} \left[\frac{\beta * N}{i}\right]$$

Where $N_c$ is the total clones number, and *n* is the selected antibodies number.

- **Affinity:** is a measure of similarity between two binary strings of equal length. The distance most often used is the Hamming distance. This distance is a simple measure that counts the number of differences of points between two strings, it is calculated as follows:

$$D = \sum_{i=1}^{L} \delta \; where \; \delta = \begin{cases} 1 & ab_i \neq ag_i \\ 0 & ab_i \equiv ag_i \end{cases}$$

Where $D$ is the Hamming distance, $ab$ is the antibody, $ag$ is the antigen and $L$ is the length of the binary string.

## IV. Results

Classification is a central problem in the pattern recognition. For each of the encountered problems (**classification**, interpretation, segmentation, …), we should detect certain characteristics to extract exploitable information in the future.

The classification improves significantly when the input vector of connectionist network includes not only the three components of the image pixels in the selected color space (RGB), but also the average and standard deviation of these components immediate vicinity of the pixel to be classified, thereby improving sampling.

The quality of the classification is evaluated using three parameters: the error, the confusion rate and the classification rate.

### A. Description of the experimental data

The remote sensing image used is a multispectral image, 7-band, acquired by the Landsat 5 TM satellite, with a 30 * 30 resolution. The study scene represents the west area of Oran city in Algeria, composed in twelve classes (see Fig. 6):

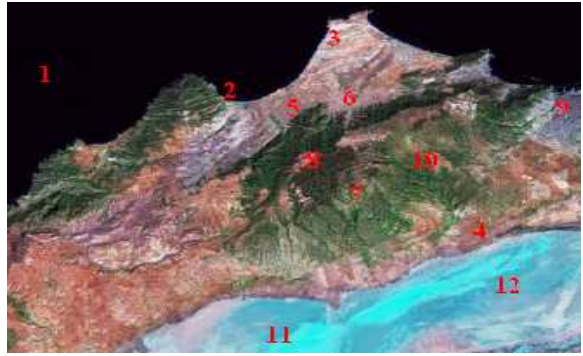

Fig. 6. Classes of the studied image

**(1)**: Sea, **(2)**: Surf, **(3)**: Sand and bare soil, **(4)**: Truck farming, **(5)**: Cultivations of cereals, **(6)**: Fallows, **(7)**: Forest, **(8)**: Scrub, **(9)**: Urban, **(10)**: Burnt land, **(11)**: Sabkha 1and **(12)**: Sabkha 2.

### B. Settings algorithms

#### 1) CLONALG :

- **Parameters to fix:** *NbrLar*: number of pixels (antibodies) of low affinity replaced, *NbrLar* = 0, *NbrTra*: number of pixels trained for each sample, *NbrTra* = 21, *Coe*: coefficient of cloning: *Coe* = 10.
- **Parameters to vary:** *Gen*: number of generations, *Ncm*: number of pixels of better affinity selected for cloning and mutation.

#### 2) API :

- **Parameters to fix:** *NbrNeur*: number of neurons in the hidden layer, *Auto*: the learning automatically stops when the results stabilize.
- **Parameters to vary:** *NbrAnt*: number of ants, $P_{local}$: patience of ants, *NbrSit*: number of hunting sites, *NbrItr*: number of iterations.

### C. Tests on AntImuClass

The performed tests are aimed to evaluate the quality of the AntImuClass network, which hybrid CLONALG and API algorithms - in the learning and the optimization phases of the RBFU network - and to test the influence of various parameters on the satellite images classification.

#### 1) Tests on CLONALG

To test the performance of the CLONALG algorithm, we vary one parameter at a time while the rest of parameters are fixed, for study the influence of each one on the neural network and thus the influence on the classification rate and therefore on the satellite images classification. The result of each test is given by the corresponding figure.

During this first test, we fix the API parameters as follows: ($A_{site}$, $A_{local}$, $P_{local}$, *NbrNeur*, *NbrSit*, *NbrItr*, *NbrAnt*) = {0.1, 0.01, 15, 12, 12, 100, 24}.

**Note:** we decided to take a number of iterations equal to 100. At first seen, this number seems fairly important and this will we avoid the above we focus in the sense that by taking it and it will not affect the results throughout the tests, as we shall see later.

*a) Influence of the number of generations*

For this test we fix the number of pixels of better affinity selected for cloning and mutation (***Ncm***) to 10, and we vary the number of generations (***Gen***). The test results are illustrated by the following table:

TABLE I.  INFLUENCE OF THE NUMBER OF GENRATIONS ON THE CLASSIFICATION RATE

| *Gen* | **Classification rate** |
|---|---|
| 10 | 93.25 % |
| 20 | **95.07 %** |
| 30 | **95.07 %** |
| 40 | 95.01 % |

Based on the above table, we find that when the number of generations is at its lowest (***Gen*** = 1), we obtain an unsatisfactory classification rate (equal to 87.23%), but when we increase the number of generations, the classification rate increases soon. With 20 generations, we note stabilization in the rate and after 30 generations a decrease in the rate was noticed.

With a number of generations equal to 20 and 30 generations, we obtain a classification rate equal to 95.07%, and even if a priori we should choose the number of 20 generations because of time savings (cause more we increase the calculation more processing time increases), we choose for the rest of tests the number for 30 generations, since, with this number we have less confusion between classes. The following figures show respectively the results of the classification made according to two different numbers of generations (see fig.7 and fig. 9).

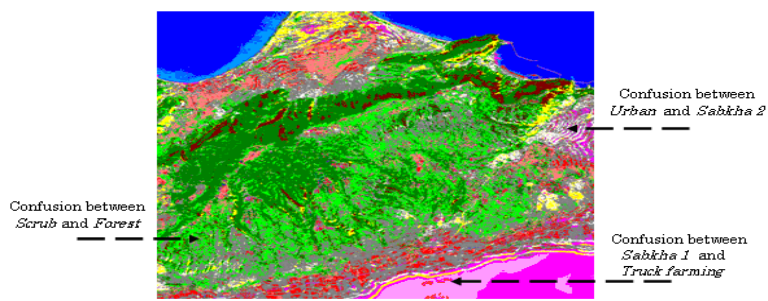

Fig. 7.  Classified image with ***Gen*** = 20, rate= 95.07%

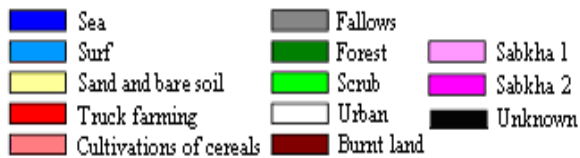

Fig. 8.  Legend of the classified images

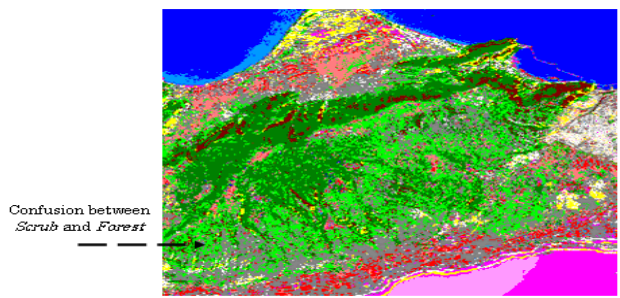

Fig. 9.  Classified image with ***Gen*** = 30, rate= 95.07%

Unlike the previous figure, with adding 10 generations, the algorithm affects all pixels of *Sabkha 1* and *2* classes to the right class. 10 generations seem a small number at first but in the end were used to a much better affectation of pixels in the right class. And like the previous classification, the confusion between the two classes *Scrub* and *Forest* persists.

*b)* *Influence of the number of pixels of better affinities*

Having achieved the maximum classification rate with the generations number variation, we thought of other changes including the number of pixels for better affinities. Now, we fix the number of generations *Gen* to 30 generations and we vary the number of pixels of better affinity selected for cloning and mutation (*Ncm*). The results are shown in the following table:

TABLE II. INFLUENCE OF THE NUMBER OF PIXELS OF BETTER AFFINITY ON THE CLASSIFICATION RATE

| *Ncm* | **Classification rate** |
|---|---|
| 5 | 92.50 % |
| 10 | **95.27 %** |
| 15 | 94.54 % |
| 20 | 93.03 % |
| 35 | 93.44% |

We note that with a number of pixels of better affinity selected for cloning (*Ncm*) equal to 10 the classification rate is at its maximum.

The following figure shows the satellite image resulted with a classification rate equal to 94.54% (see Fig. 10), and where we see that the number of confusions in this figure greatly increased compared to that where the classification rate is equal to 95.27% (Figure 11).

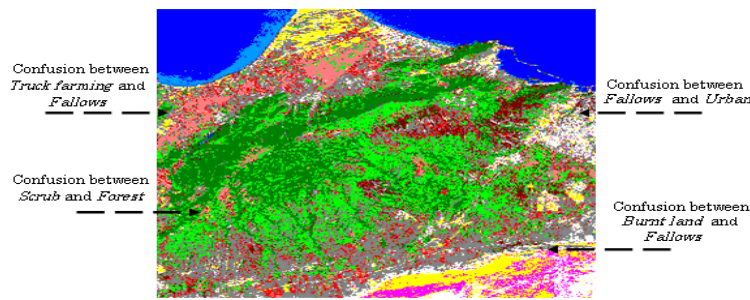

Fig. 10. Classified image with rate = 94.54 %

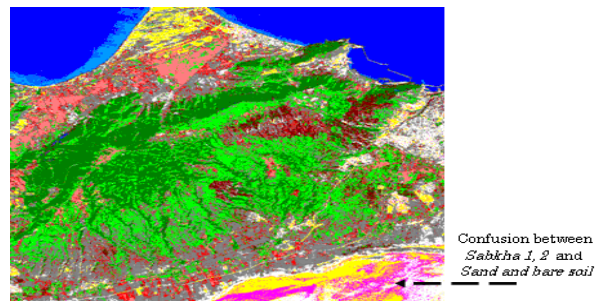

Fig. 11. Classified image with rate= 95.27 %

In the figure 12, even if the classification rate is quite important but it's the confusion between *Sabkha 1*, *Sabkha 2* and *Sand and bare soil* stands which arises the most.

Consequently, with 10 pixels of better affinities the network offers the best result of classification.

*2) Test on API*

From the previous tests, it seems clear that with the CLONALG parameters equal to 10 for *Ncm* and to 30 for *Gen*, we get the best results.

In the following and throughout tests, we fix the parameters of the CLONALG algorithm, i.e.: (***NbrLar***, ***NbrTra***, ***Coe***, ***Gen***, ***Ncm***) respectfully as follows = {0, 21, 10, 30, 10}.

*a) Influence of the ants number*

For seeing if the number of ants has an impact on the satellite images classification and for determinate the degree of this impact, we vary the number of ants while the others algorithm parameters are fixed, in advance: ($A_{site}$, $A_{local}$, $P_{local}$, *NbrNeur*, *NbrSit*, *NbrItr*) to the following values (0.1, 0.01, 15, 12, 12, 35).

TABLE III.  INFLUENCE OF THE NUMBER OF ANTS ON THE CLASSIFICATION RATE

| *NbrAnt* | **RBFU-API rate** | **RBFU-API$_h$ rate** |
|---|---|---|
| 5 | 90.61 % | 91.36 % |
| 10 | 89.95 % | 93.09 % |
| 20 | **90.31 %** | 94.54 % |
| 24 | 90.23 % | **95.47 %** |
| 30 | 90.03 % | 95.25 % |
| 35 | 87.45 % | 93.36 % |

The obtained results throughout this test show that the rates are not stable, since an increase of the number of ants doesn't necessarily mean an increased in the classification rate. However it's with a number of ants equal to 20 that the classification rate (RBF-API) is at its maximum.

At this stage we introduce the factor of heterogeneity to the RBF-API network and thus all levels of this test are increased, all without exception. However it is not with a number of ants equal to 20 that we get this time the highest rate but with a number equal to 24.

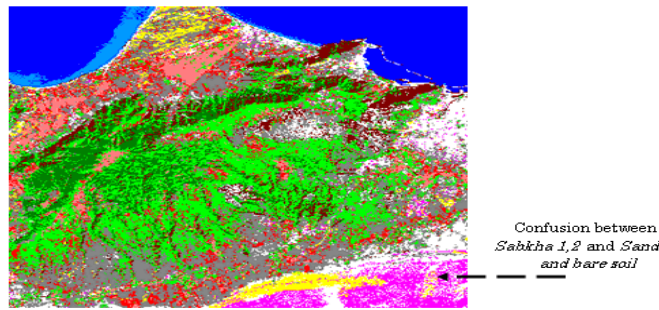

Fig. 12.  Classified image with a rate = 94.67 %

*b) Influence of the patience of ants*

The influence of the patience of ants on the classification rate of satellite images is described in the following table, where we vary the patience of ants.

TABLE IV.  INFLUENCE OF PATIENT ANTS ON THE CLASSFICATION RATE

| $P_{local}$ | **RBFU-API rate** | **RBFU-API$_h$ rate** |
|---|---|---|
| 2 | 90.04 % | 90.46 % |
| 5 | 89.25 % | 93.96 % |
| 15 | 91.77 % | **96.19 %** |
| 20 | 90.36 % | 93.63 % |
| 30 | **92.24 %** | 94.28 % |
| 35 | 90.65 % | 91.55 % |
| 50 | 92.02 % | 95.17 % |

Again the best results were obtained at different levels ($P_{local}$ is equal to 30 for RBF-API and to 15 for RBF-API$_h$).

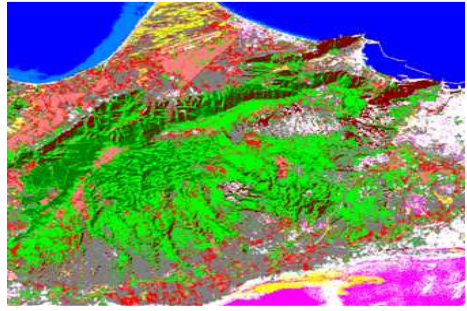

Fig. 13.  Classified image with a rate = 92.24 %

*c)* *Influence of the number of sites for hunting*

As done previously, we fix all the algorithm parameters unless the number of sites that we vary and we note the changes as follows.

TABLE V.   INFLUENCE OF THE NUMBER OF SITES ON THE CLASSIFICATION RATE

| *NbrSit* | **RBFU-API rate** | **RBFU-API$_h$ rate** |
|---|---|---|
| 5 | 54.85 % | 62.65 % |
| 10 | 87.33 % | 87.93 % |
| 12 | **89.74 %** | **96.87 %** |
| 20 | 87.55 % | 87.99 % |
| 35 | 83.15 % | 88.23 % |
| 50 | 77.56 % | 89.41 % |

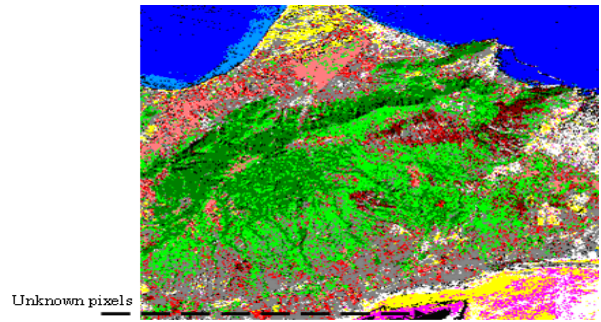

Fig. 14.  Classified image with a rate = 89.74 %

Unlike other figures, in Figure 10 we note that many pixels were not classified, even worse, they were detected as unknown pixels, this means that the algorithm has failed in what class (category) put them.

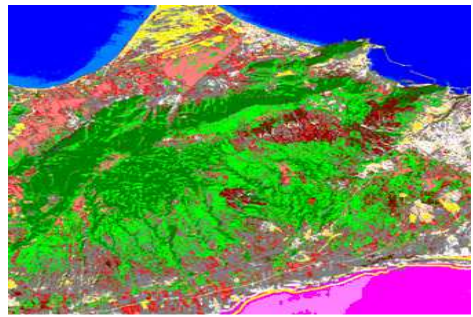

Fig. 15.  Classified image with a rate = 96.87 %

For the first time since the early tests, the best results were obtained at the same level, whether for a homogeneous population (API) or heterogeneous (API$_h$), and that for a number of sites equal to 12. However, it is more the API$_h$ result that interests us because it is through this diversity that we increase the quality of results.

### d) *Influence of number of iterations*

We vary the number of iterations and note the changes.

TABLE VI. INFLUENCE OF THE NUMBER OF ITERATIONS ON THE CLASSIFICATION RATE

| *NbrItr* | **RBFU-API rate** | **RBFU-API$_h$ rate** |
|---|---|---|
| 15 | 89.63 % | 92.89 % |
| 25 | 91.28 % | 94.93 % |
| 30 | 91.78 % | 95.07 % |
| 35 | **92.00 %** | **97.02 %** |
| 40 | 92.00 % | 95.26 % |

We note that it isn't necessary to go up to 50 iterations, because with only 35 iterations we obtain the same results.

As with the previous test best results were obtained at the same level, 35 iterations, in addition to this the more we increase the number of iterations over the classification rate increases.

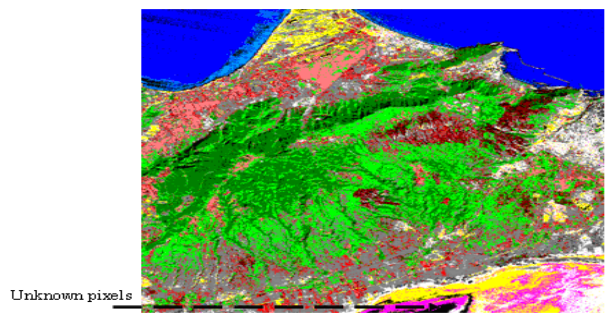

Fig. 16. Classified image with rate = 92.00 %

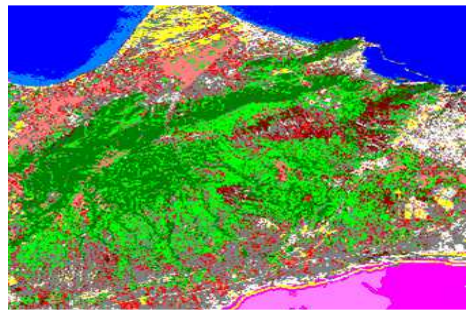

Fig. 17. Classified image with rate = 97.02 %

In conclusion, AntImuClass performs better than RBF alone [6] but requires a longer execution time. We note also that the time limit given to our algorithms was fixed relative to ACO algorithms that, unlike the local search algorithm converge slowly enough GNGU [6] has been disadvantaged by this time limit.

Finally, the parameters of the AntImuClass algorithm - CLONALG and API combined - offer the best results are equivalent to: (***NbrNeur, NbrAnt, NbrSit, P$_{local}$, NbrItr, NbrLar, NbrTra, Coe, Gen, Ncm***) = {12, 24, 15, 12, 35, 0, 21, 10, 30, 10}.

## V. CONCLUSION

We were interested in this paper on problem of the remote sensing imagery classification, through a hybrid network type, named AntImuClass: RBF-GNGU network hybridized, based on an optimization by clonal selection (immune system) doped with a learning based on colony ant by *Pachycondyla apicalis* type.

This algorithm was tested on satellite images from Landsat 5 TM satellite of the western region of Oran city, where it has shown promising results. Thus, a natural biological mechanism of self-government of autonomous agents can manage complex problems in real time with great efficiency.

The presented model is quite flexible, it may incorporate several variations. One can, for example, consider that the environment contains many sources of food in each set amount, which decreases with the visits of ants. These quantities can be incorporated into the model of the optimal control function through the award states sources. The reward then evolves over time, and traces of pheromones, which are constantly updated, follow this optimal value function "change". You can also use different settings for the outward and return journey of ants: this could model different strategies depending on whether or not the ants carry food.

In conclusion, we showed that once again the bio-inspired computing demonstrates its effectiveness in the face of problems like classification, and is able to classify a relevant data bases.

* Y.T Amghar, Science and Techniques Preparatory School of Oran, ALGERIA
*E-mail address*: amgharyasmina@yahoo.fr

[†] H. Fizazi, University of Science and Technology of Oran, ALGERIA
*E-mail address*: fizazihadria@yahoo.fr